\documentclass[twoside]{article}

\usepackage[accepted]{aistats2020}
\usepackage{amssymb}
\usepackage{xcolor}
\usepackage{algorithm}
\usepackage{algorithmic}
\usepackage{graphicx}
\usepackage{tabularx}
\usepackage{booktabs} 
\usepackage{multirow}

\newcommand{\KL}{D_{\mathrm{KL}}}

\usepackage{amsmath,amsfonts,bm,amssymb}

\newtheorem{proposition}{Proposition}

\def\eqref#1{equation~\ref{#1}}

\def\1{\bm{1}}

\DeclareMathAlphabet{\mathsfit}{\encodingdefault}{\sfdefault}{m}{sl}
\SetMathAlphabet{\mathsfit}{bold}{\encodingdefault}{\sfdefault}{bx}{n}

\newcommand\blfootnote[1]{%
  \begingroup
  \renewcommand\thefootnote{}\footnote{#1}%
  \addtocounter{footnote}{-1}%
  \endgroup
}

\usepackage[round]{natbib}

\begin{document}

%

%
\runningauthor{ Yuxuan Song, Ning Miao, Hao Zhou, Lantao Yu, Mingxuan Wang, Lei Li}

\twocolumn[

\aistatstitle{Improving Maximum Likelihood Training for Text Generation with Density Ratio Estimation}

\aistatsauthor{ Yuxuan Song \And Ning Miao  \And Hao Zhou \And Lantao Yu}

\aistatsaddress{Shanghai Jiao Tong University  \And Bytedance AI lab
\And Bytedance AI lab \And Stanford University}
\aistatsauthor{ Mingxuan Wang \And Lei Li}
\aistatsaddress{ Bytedance  AI lab \And Bytedance  AI lab}
]




\begin{abstract}

Autoregressive sequence generative models trained by Maximum Likelihood Estimation suffer the exposure bias problem in practical \emph{finite} sample scenarios. 
The crux is that the number of training samples for Maximum Likelihood Estimation is usually limited and the input data distributions are different at training and inference stages.
Many methods have been proposed to solve the above problem \citep{yu2017seqgan,lu2018cot}, which relies on sampling from the non-stationary model distribution and suffers from high variance or biased estimations.
In this paper, we propose $\psi$-MLE, a new training scheme for autoregressive sequence generative models, which is effective and stable when operating at large sample space encountered in text generation. 
We derive our algorithm from a new perspective of self-augmentation and introduce bias correction with density ratio estimation.
Extensive experimental results on synthetic data and real-world text generation tasks demonstrate that our method stably outperforms Maximum Likelihood Estimation and other state-of-the-art sequence generative models in terms of both quality and diversity.
\end{abstract}
\section{Introduction}
Deep generative models dedicate to learning a target distribution and have shown great promise in numerous scenarios, such as image generation  \citep{arjovsky2017wasserstein,goodfellow2014generative}, density estimation  \citep{ho2019flow++,salimans2017pixelcnn++,kingma2013auto,townsend2019practical}, stylization \citep{ulyanov2016instance}, and text generation  \citep{yu2017seqgan,li2016deep}.
Learning generative models for text data is an important task which has significant impact on several real world applications, \emph{e.g.}, machine translation, literary creation and article summarization. However, text generation remains a challenging task due to the discrete nature of the data and the huge sample space which increases exponentially with the sentence length. 


Text generation is nontrivial for its huge sample space. For generating sentences of various lengths, current text generation models are mainly based on density factorization instead of directly modeling the joint distribution, which results in the prosperity of  neural autoregressive models on language modeling. As neural autoregressive models have explicit likelihood function, it is straightforward to employ Maximum Likelihood Estimation (MLE) for training. 
Although MLE is is asymptotically consistent, for practical \textbf{finite} sample scenarios, it is prone to overfit on the training set.
Additionally, during the inference (generation) stage, the error at each time step will accumulate along the sentence generation process, which is also known as the \emph{exposure bias}  \citep{ranzato2015sequence} problem.
\blfootnote{The work was done during the first author's internship  at Bytedance AI lab.}

Many efforts have been devoted to address the above limitations of MLE. Researchers have proposed several non-MLE methods based on minimizing different discrepancy measures, \emph{e.g.}, Sequential GANs  \citep{yu2017seqgan,che2017maximum,gumbelgan} and CoT \citep{lu2018cot}. 
However, non-MLE methods typically relies on sampling from the generative distribution to estimate gradients, which results in high variance and instability during training, as the generative distribution is non-stationary during training process. 
Some recent study \citep{caccia2018language} empirically shows that  non-MLE methods potentially suffer from mode collapse problem and cannot actually outperform MLE in terms of quality and diversity tradeoff. 

In this paper, we seek to leverage the ability of generative models itself for providing unlimited amount of samples to augment the training dataset, which has the potential of alleviating the overfitting problem due to limited samples, as well as addressing the exposure bias problem by providing the model with prefixes (input partial sequences) sampled from its own distribution. 
To correct the bias incurred by sampling from the model distribution, we propose to learn a progressive density ratio estimator based on Bregman divergence minimization. The above procedures together form a novel training scheme for sequence generative models, termed $\psi$-MLE.

Another essential difference between MLE and $\psi$-MLE lies in the fact that the likelihood of samples not in training set are equally penalized through normalization in MLE, whether near or far from the true distribution. While $\psi$-MLE takes the difference in the quality of unseen samples into account through the importance weight assigned by density ratio estimator, which can be expected to get further improvement.

Empirically, MLE with mixture training data gives the same performance as vanilla MLE training with only training data. 
But our proposed $\psi$-MLE consistently outperforms vanilla MLE training.
Additionally, we empirically demonstrate the superiority of our algorithm over many strong baselines like GAN in terms of generative performance (in the quality-diversity space) with both synthetic and real-world datasets. 

\section{Preliminary} 
\subsection{Notations}

We denote the target data distribution as $p_{\text {data }}$, and the empirical data distribution as $\hat{p}_{\text {data }}$. The parameters of the generative model $G$ are presented by $\theta$ and the parameters of a density ratio estimator $r$ are presented by $\psi$. $p_\theta$ denotes the distribution implied by the tractable density generative model $G$. The objective is to fit the underlying data distribution $p_\text{data}$ with a parameterized model distribution $p_\theta$ with empirical samples from $p_\text{data}$. We use $s$ to stand for a sample sequence from datasets or from generator's output. And $s_l$ stands for the l-th token of $s$, where $s_0=\emptyset$.  

\subsection{MLE  vs Sequential GANs}
\label{intro_mle}
It should be noticed that both MLE and GANs for sequence generation suffer from their corresponding issues. 
In this section, we delve deeply into the specific properties of MLE and GANs,
and explore how these properties affect their performances in modeling sequential data.

\paragraph{MLE} The objective of  Maximum Likelihood Estimation (MLE) is:
\begin{equation}
\label{mle}
L_{\mathrm{MLE}}(\theta) = \mathbb{E}_{s \sim p_{\text {data }}}\left[\log p_{\theta}(s)\right]
\end{equation}
where $p_{\theta}(s)$ is the learned probability of sequence $s$ in the generative model.
Maximizing the objective is equivalent to minimizing the Kullback-Leibler (KL) divergence:
\begin{equation}
\KL(p_{\text {data }}|| p_\theta)=\mathbb{E}_{s \sim p_{\text {data }}} \log \frac{p_{\text {data }}(s)}{p_\theta(s)}
\end{equation}
Though MLE has lots of attractive properties, it has two critical issues:

1)~MLE is prone to overfitting on small training sets. 
Training an autoregressive sequence generative model with MLE on a training set consists of sentences of length $L$, the standard objective can be derived as following:
\begin{equation}
\label{teacher forcing}
L_{\hat{\mathrm{MLE}}}(\theta)=\underset{s \sim \hat{p_{\text{data}}}}{\mathbb{E}} \sum_{l=1}^{L} \log p_{\theta}\left(s_{l} | s_{1 : l-1}\right)
\end{equation}
The forced exposure to the ground-truth data shown in Eq.~\ref{teacher forcing} is known as ``teacher forcing", which causes the problem of overfitting.
What makes thing worse is the exposure bias. During training, the model only learns to predict $s_l$ given $s_{1:l-1}$, which are fluent prefixes in the training set. During sampling, when there are some small mistakes and the first $l-1$ can no longer make up a very fluent sentence, the model may easily fail to predict $s_l$.

2)~KL-divergence punishes the situation where the generation model gives real data points low probabilities much more severely than that where unreasonable data points are given high probabilities. As a result, models trained with MLE will focus more on not missing real data points than avoiding generating data points of low quality.

\paragraph{Sequential GANs}
 Sequential GANs~\citep{yu2017seqgan,guo2018long}, are proposed to overcome the above shortcomings of MLE.  The typical objective of them is:
\begin{equation}
\label{lgan}
\begin{aligned}
&L_{\mathrm{GAN}}(\theta) = \\
&\min _{\theta}-\mathbb{E}_{s \sim p_{\theta}}\left[\sum_{t=1}^{n} Q_{t}\left(s_{1:t-1}, s_{t}\right) \cdot \log p_{\theta}\left(s_{t} | s_{1:t-1}\right)\right]
\end{aligned}
\end{equation}
$Q_{t}\left(s_{1:t-1}, s_{t}\right)$ is action value, which is usually approximated by a discriminator‘s evaluation on the complete sequences sampled from the prefix $s_{t+1} = [s_{1:t-1}, s_{t}]$. The main advantage of GANs is that when we update the generative model, error will be explicitly reduced by the effect of normalizing constant. 

However, there is also a major drawback of GANs. As the gradient is estimated by REINFORCE algorithm \citep{yu2017seqgan}, the generated distribution is non-stationary. As a result, the estimated gradient may suffer from high variance. Though many methods have been proposed to stabilize the training of sequential GANs, \emph{e.g} control variate \citep{che2017maximum} or MLE pretraining \citep{yu2017seqgan}, there, they only have limited effect on sequential data. Moreover, as indicated by recent works \citep{caccia2018language}, sequential GANs sharpen density functions in the distribution's support, which sacrifices diversity for better quality.

\section{Methodology}
In order to combine the advantages of MLE, which directly trains the model on high-quality training samples, and GANs, which actively explore unseen spaces, we propose $\psi$-MLE. We further remove noise points of $\psi$-MLE by performing importance sampling whose weight is given by a density ratio estimator. 

\subsection{$\psi$-MLE for Sequence Generation}

The different properties of MLE and GANs mainly result from $O$, the effect zone of supervision. To be concrete, $O$ is a subset of all possible data points, whose likelihoods are directly updated during training.
MLE only maximizes the probabilities of points in the training set, which is discrete and finite. However, the actual data space contains far more points than the training set, on which there is no supervision. In contrast, as the generators of GANs are able to generate all possible data points, $O_{GAN}$ is essentially the whole data space. Large enough though $O_{GAN}$ is, the supervision signal, i.e., the gradients for updating GANs' generators usually have high variances compared with the gradients of MLE.

To combine the merits of both methods, we propose $\psi$-MLE which blends samples generated by the current generation model into training data:
\begin{equation}
p_{\operatorname{mix}}(S)=m p_{\mathrm{data}}(S)+(1-m) p_{\theta}(S),
\label{eqn:m}
\end{equation}
where $m \in \left[0,1\right]$ is the proportion of training data. 
By $\psi$-MLE, we extend $O$ to the whole space. And since there are real training data in the mixture samples, the gradients are more informative with lower variances.

For training, we directly minimize the forward KL divergence between  $p_{\operatorname{mix}}$ and $p_\theta$, which is equivalent to performing MLE on samples from $p_{\operatorname{mix}}$. Since the training goal at each step is to maximize:
\begin{equation}
\label{Reweighted_target}
    \mathrm{E}_{p_{\operatorname{mix}}(S)}[\log p_\theta(S)],
\end{equation}
when the KL-divergence decrease, the gap between $p_\theta$ and $p_{data}$ get smaller. Eventually, when $p_\theta \approx p_{\operatorname{mix}}$, $p_\theta$ also approximates $p_{data}$.


However, $p_{mix}$ may be very different from $p_{data}$, especially at the beginning of training. This discrepancy may result in generating really poor samples which have high likelihoods in $p_\theta$ but not in $p_{data}$. As a result, the training set gets noisier, which may harm performance.


\subsection{Noise Reduction by Importance Sampling}
\label{DDRE}
To make the distribution of training samples closer to $P_{data}$, we introduce the following importance sampling method. The main idea is to first get a batch of samples from $P_{mix}$, and then give each sample an importance weight $r$ according to its similarity with real samples. Then the training objective turns into:
\begin{equation}
\label{}
    \mathrm{E}_{p_{\operatorname{mix}}(S)}[r_\psi(S)\log p_\theta(S)],
\end{equation}
where $\psi$ is the parameter of the importance weight estimator.


In ideal conditions, where $r_\psi(S)=r_{\text {optimal}}(S)=\frac{p_{\text {data }}(S)}{p_{\operatorname{mix}}(S)}$, the training essentially minimizes the KL-divergence between $P_\theta$ and real data distribution $P_{data}$:
\begin{equation}
\label{bias_correction}
\begin{aligned}
&\mathrm{E}_{p_{\operatorname{mix}}(S)}[r_\psi(S)\log p_\theta(S)]
\\=&\mathrm{E}_{p_{\operatorname{mix}}(S)}[\frac{p_{\text {data }}(S)}{p_{\operatorname{mix}}(S)}\log p_\theta(S)] \\
=&\mathrm {E}_{p_{\text {data }}(S)}[\log p_\theta(S)]
\\
 \approx &\frac{1}{T} \sum_{i=1}^{T} \frac{p_{\text {data }}\left(s^{(i)}\right)}{p_{\operatorname{mix}}\left(s^{(i)}\right)} \log p_\theta\left(s^{(i)}\right), 
\end{aligned}
\end{equation}
where s in the last equation are samples from $P_{mix}$.
We assert that dividing $p_{\text {data }}$ by $p_{\operatorname{mix}}$ won't cause any numerical problem, since the support of $p_{\text {data }}$ is a subset of $p_{\operatorname{mix}}$'s support.
However, it is infeasible to directly calculate $r_{\text {optimal}}=\frac{p_{\text {data}}} {p_{\operatorname{mix}}}$. So we need to approximate it by $r_\psi$.
The first thought is to use a new parametric model $p_{\beta}$ to approximate $p_{\text {mix}}$, and set $r_\psi=1-\frac{p_{\theta}}{p_{\beta}}$. But this method will lead to severe numerical instability. In this paper, we choose to directly approximate $r_{\text {optimal}}$ by training a discriminator between $p_{\text {data}}$ and $p_{\text {mix}}$.
To be more concrete, we first assign positive labels $y=1$ to samples from $p_{\text {data}}(s)$ and negative labels $y=0$ to samples from $p_{\operatorname{mix}}(s)$. Then we train a probabilistic classifier $c$: $\mathcal{S} \rightarrow [0,1]$ to output the probability of $s$ belonging to each class. After the training of $c$ converges, we set $r_\psi=\gamma \frac{c(\mathbf{s})}{1-c(\mathbf{s})}$ and get the following proposition:
\begin{proposition}
With a Bayes optimal classifier $c$,
\begin{equation}
r_\psi=r_{\text{optimal}}.
\end{equation}
\end{proposition}
$\gamma = \frac{p(y=0)}{p(y=1)}$ 
is the amount ratio of negative samples and positive samples.
We keep $\gamma=1$ by using the same number of negative and positive samples in a mini-batch.

Note that the density ratio is obtained indirectly from the classifier $c$ which is typically poorly calibrated. Therefore we need to frequently calibrate $c$ to get a better density ratio estimation and avoid numerical problems caused by miscalibration~\citep{turner2018metropolis}.
However, it can be quite computationally expensive to calibrate $c$ after each update. To sidestep the above obstacle, directly estimating $\frac{p_{\text {data }}(s)}{p_{\operatorname{mix}}(s)}$ is a more general approach, which may lead to a more accurate density ratio estimation than the ``classifier-based" method mentioned above.

Given two distributions $p(x)$ and $q(x)$, the target of direct density estimation is to obtain a density ratio model $r_\psi(x)$, which can directly approximate the true density ratio $r(x) = \frac{p(x)}{q(x)}$.  \citep{sugiyama2012density,uehara2016generative} proposed to utilize the Bregman divergence 
as a measure of the discrepancy between two density ratio functions, which guides the training of density ratio model. The Bregman divergence is an extension of Euclidean distance which measures the distance between two data points $x_1$ and $x_2$, and the definition with respect to function $f$ is as following:
\begin{equation}
B R^{\prime}_{f}\left(x_1 \| x_2\right)=f\left(x_1\right)-f(x_2)-\nabla f(x_2)\left(x_1-x_2\right)
\end{equation}
where $f: \Omega \rightarrow \mathcal{R}$ is a strictly convex and continuously differentiable function defined on a closed set $\Omega$. 

The integration of Bregman divergence $B R^{\prime}_{f}\left[r(x) \| r_{\theta}(x)\right]$ between an estimated density ratio function $r_\psi(x)$ and the real density ratio function $r(x)$ under measure $q(x)dx$ is as following:
\begin{equation}
\label{b_divergence}
\begin{aligned} B R_{f}(r \| r_{\psi}) =&\int B R^{\prime}_{f}[r(x) \| r_{\psi}(x)] q(x) d x \\ =&\int (f(r(x))-f(r_{\psi}(x))-\\
&\nabla f(r_{\psi}(x))(r(x)-r_{\psi}(x))) q(x) d x.
\end{aligned}
\end{equation}
Then the estimation procedure can be turned into an optimization procedure with respect to the parameter $\psi$. We leave the discussion with different selections of $f$ in Sec.~\ref{relation_fd}.
In practical training, we alternatively update $\psi$ and $\theta$. The whole training procedure is in Algorithm~\ref{alg:pro_correction}.
\begin{algorithm}[t]
  \caption{Progressive Bias Correction}
  \label{alg:pro_correction}
  \begin{algorithmic}[1]
  \STATE {\bfseries Require:} Generator $p_\theta$ with parameter as $\theta$; Density Ratio estimator $r_\psi$ with parameter as $\psi$; Empirical data distribution as $\hat{p_{\text{data}}}$; A mixture weight $m$.
  \REPEAT
  \STATE Sample two minibatches of samples $\left\{x_{1}, \dots, x_{B}\right\}$, $\left\{x_{B+1}, \dots, x_{2B}\right\}$ from $p_\theta$.
  \STATE Sample a minibatch of samples $\left\{y_{1}, \dots, y_{B}\right\}$ from $\hat{p_{\text{data}}}$
  \STATE Create a mixed minibatch $\left\{z_{1}, \dots, z_{B}\right\}$ by mixing samples from $\left\{x_{B+1}, \dots, x_{2B}\right\}$ and $\left\{y_{1}, \dots, y_{B}\right\}$ according to the mixture weight $m$.
  \FOR{number of $\psi$ update}
  \STATE Update $\psi$ according to Eq.~\ref{b_divergence}:

 \STATE $\psi^{t+1}=\psi^{t}-\nabla_{\psi}\frac{1}{B} \sum_{i=1}^{B} (\nabla f\left(r_{\psi}\left(x_{i}\right)\right) r_{\psi}\left(x_{i}\right)-f\left(r_{\psi}\left(x_{i}\right)\right)-\nabla f\left(r_{\psi}\left(y_{i}\right)\right))$
  \ENDFOR
  \FOR{number of $\theta$ update}
  \STATE Update $\theta$ according to Eq.~\ref{Reweighted_target}:
  \STATE $\theta^{t+1} = \theta^{t} - \frac{1}{B} \sum_{i=1}^{B}(r_\psi(z_i)\nabla_{\theta}\log p_\theta(z_i)) $
  \ENDFOR
  
  \UNTIL{Convergence}
  \STATE {\bfseries Output:} 
  \end{algorithmic}
\end{algorithm}

\section{Connection with other methods}

In this section, we provide further investigation on direct density ratio estimation and theoretical justification for our proposed methods.


\subsection{Relation with GANs}
As introduced in Sec.~\ref{intro_mle}, sequential GANs usually adopt policy gradient methods for training. Their objectives can be interpreted in a Reinforcement Learning(RL) fashion:
\begin{equation}
\begin{aligned}
\mathcal{L}_{\text{RL-GAN}}(\boldsymbol{\theta} ; \tau, \hat{p}_{\text{data}})= &-\tau \mathbb{H}\left(p_{\theta}(s)\right) \\
&-\sum_{s \in \mathcal{S}} p_{\theta}(s) r\left(s, \hat{p}_{\text{data}}\right).
\end{aligned}
\end{equation}
In this formula, $r\left(s, \hat{p}_{\text{data}}\right)$ is the reward function, which is usually implemented by a discriminator. In order to mitigate mode collapse,  the regulation term $\mathbb{H}\left(p_{\theta}(s)\right)$ is added. We further introduce an \emph{exponentiated payoff distribution}~\citep{norouzi2016reward}
\begin{equation}
q\left(s ; \tau\right)=\frac{1}{Z\left(s, \tau\right)} \exp \left\{r\left(s, \hat{p}_{\text{data}}\right) / \tau\right\}
\end{equation}
Then, we can see that training discrete GANs are essentially minimizing the following KL divergence, $\KL(p_{\theta}(s)||q\left(s ; \tau\right))$, which is shown at following:
\begin{align}
    &\KL(p_{\theta}(s)||q\left(s ; \tau\right)) \nonumber \\ 
    &=\mathbb{E}_{s \sim p_\theta}(\log p_\theta(s)-\log(\frac{1}{Z(s, \tau)} \exp \{r(s, \hat{p}_{\text{data}}) / \tau\})) \nonumber\\
    & = \frac{1}{\tau} \mathcal{L}_{\text{RL-GAN}}(\theta ; \tau)+\text { constant }.
\end{align}
The last holds by the fact that $Z(s, \tau)$ is a constant during the optimization of $\theta$.
Our method can be seen as optimizing the opposite direction of the KL divergence, \emph{i.e}, $\KL(q\left(s ; \tau\right) ||p_{\theta}(s))$. As it is intractable to directly sample from $q\left(s ; \tau\right)$, we first sample from $p_{\operatorname{mix}}$ and conduct importance sampling with weight $r_\psi$ to obtain unbiased estimation of $\KL(q\left(s ; \tau\right) ||p_{\theta}(s))$.

\subsection{Relation with $f$-Divergence}
\label{relation_fd}
Density ratio estimation is closely to related f-divergence~\citep{nowozin2016f}, which measures the difference between two probability distributions. 
Given two distributions with absolutely continuous density functions $p$ and $q$, the f-divergence is defined as:
    
\begin{equation}
    D_{f}(p \| q)=\int_{\mathcal{X}} q(x) f\left(\frac{p(x)}{q(x)}\right) \mathrm{d}x = E_{q}\left[f\left(\frac{p(x)}{q(x)}\right)\right]
\end{equation}

where $f : \mathbb{R}_{+} \rightarrow \mathbb{R}$ is a convex and lower-semicontinous function with $f(1)= 0$. 

If $f$ is a strictly convex and continuously differentiable
 function, the following conclusions can be derived. 
\begin{proposition}
Minimizing Bregman divergence between two distributions $p$ and $q$ with respect to $f$ is essentially estimating the $f$-divergence between $p$ and $q$ with $\nabla f(r_{\psi}(x))$ as the dual coordinates.
\end{proposition}

When the true density ratio are available, the f-divergence can also be obtained, it is not surprising that estimating density ratio by minimizing Bregman divergence with respect to function $f$ is essentially the dual of estimating the $f$-divergence by maximizing a variational bound. 
We rewrite the Eq.~\ref{b_divergence} as following:
\begin{equation}
\label{f_bf}
\begin{aligned} &B D_{f}(r \| r_{\psi})\\
=&\int (f(r(x))-f(r_{\psi}(x))\\
&-\nabla f(r_{\psi}(x))(r(x)-r_{\psi}(x))) q(x) d x \\
=& E_{q}\left[f\left(\frac{p(x)}{q(x)}\right)\right] -E_{x \sim p}\left[\nabla f\left(r_{\psi}(x)\right)\right] +\\ 
&E_{x \sim q}\left[\left(\nabla f\left(r_{\psi}(x)\right) r_{\psi}(x)-f\left(r_{\psi}(x)\right)\right)\right]
\end{aligned}
\end{equation}

After some simple operations, we get:
\begin{equation}
\begin{aligned}
    &E_{x \sim p}\left[\nabla f \left(r_{\psi}(x)\right)\right] + \nonumber \\
    &E_{x \sim q}\left[\left(\nabla f\left(r_{\psi}(x)\right) r_{\psi}(x)-f\left(r_{\psi}(x)\right)\right)\right]\nonumber \\
    = &E_{q}\left[f\left(\frac{p(x)}{q(x)}\right)\right] - B D_{f}(r \| r_{\psi}) \leq E_{q}\left[f\left(\frac{p(x)}{q(x)}\right)\right]
\end{aligned}
\end{equation}
The inequality holds for the fact that $B D_{f}(r \| r_{\psi}) \geq 0$, and the equality holds if and only if $r_{\psi}(x) = r(x)$. 

Meanwhile, the dual representation of $f$-divergence \citep{nowozin2016f} is illustrated as follows:
\begin{equation}
\label{fd-bd}
\begin{aligned} &D_{f}(p \| q)  \geq \sup _{T \in \mathcal{T}}\int_{\mathcal{X}} p(x) T(x) \mathrm{d} x-\int_{\mathcal{X}} q(x) f^{*}(T(x)) \mathrm{d} x \\ =&\sup _{T \in \mathcal{T}}\left(\mathbb{E}_{x \sim p}[T(x)]-\mathbb{E}_{x \sim q}\left[f^{*}(T(x))\right]\right)
\\ =&\sup _{r_\psi}(E_{x \sim p}\left[\nabla f\left(r_{\psi}(x)\right)\right] 
\\&\quad \quad \quad+ E_{x \sim q}\left[\left(\nabla f\left(r_{\psi}(x)\right) r_{\psi}(x)-f\left(r_{\psi}(x)\right)\right)\right]),
\end{aligned}
\end{equation}
where $\mathcal{T}$ is an arbitrary class of functions $T$ and $f^{*}$ corresponds to the Fenchel conjugate of $f$. Last equation in Eq.~\ref{fd-bd} is valid for the fact that $f^{*}\left(f^{*}\left(r_{\theta}(x)\right)\right)=f\left(r_{\theta}(x)\right) r_{\theta}(x)-f\left(r_{\theta}(x)\right)$. Above discussions also indicate the knowledge distillation perspective of our method, \emph{i.e.}, we minimize the $f$-divergence between $r_\psi p_{\operatorname{mix}}$ and $p_{\mathrm{data}}$ and then distill the knowledge  by minimizing the KL divergence between  $r_\psi p_{\operatorname{mix}}$ and $p_\theta$.

\section{Experiments}
\label{fg:TEMP}
\begin{figure}[!ht]
\centering
\includegraphics[width=0.6\linewidth]{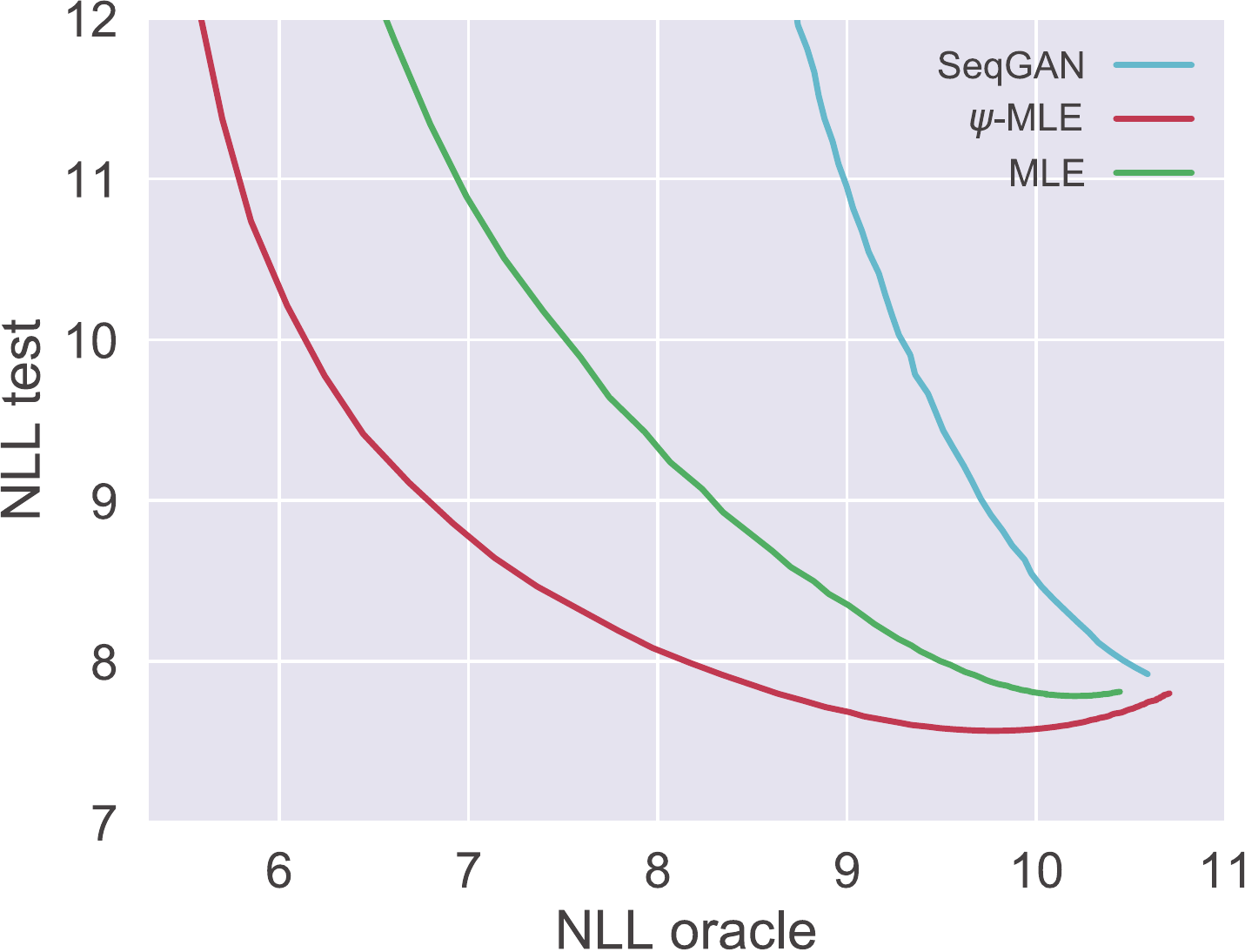}
\caption{temperature curve for MLE, $\psi$-MLE and SeqGAN}
\label{fig:model_temp}
\end{figure}

\begin{table*}[!ht]
  \caption{Likelihood-based benchmark and time statistics for synthetic Turing test within the temperature scope~\citep{caccia2018language}.}
  \label{tb:synthetic_data}
  \begin{center}
  \begin{small}
  \begin{tabular}{lccc}
    \toprule
    Model     & NLL$_{oracle}$     & NLL$_{test}$ & best NLL$_{oracle}$ + NLL$_{test}$ \\
    \midrule
    MLE & 5.53 & 7.58 & 16.28\\
    SeqGAN \citep{yu2017seqgan} & 8.12 & 7.92 & 18.44 \\
    COT \citep{lu2018cot}&  6.20 & \textbf{7.56} & 16.32\\
    LeakGAN \citep{guo2018long}&  10.01 & 8.52 & 19.45 \\
    \cmidrule{1-4}
    $\psi$-MLE & \textbf{5.09} & \textbf{7.56} & \textbf{15.98}\\
    \bottomrule
    \end{tabular}
    \end{small}
    \end{center} 
\end{table*}


\begin{table*}[!ht]
\centering
\label{tb:emnlp}
\caption{Test BLEU and Self-BLEU on EMNLPNEWS.}
\begin{center}
\begin{small}
\begin{tabular}{l cccc | cccc }
\toprule
 & \multicolumn{4}{c|}{BLEU($\uparrow$)} & \multicolumn{4}{c}{Self-BLEU($\downarrow$)}\\\hline
 & 2 & 3 & 4 & 5 & 2 & 3 & 4 & 5 \\
\midrule
Training Data & 0.86 & 0.61 & 0.38 & 0.23 & 0.86 & 0.62 & 0.38 & 0.24 \\
SeqGAN \citep{yu2017seqgan} & 0.72 & 0.42 & 0.18 & 0.09  & 0.91 & \textbf{0.70} & \textbf{0.46} & 0.27 \\
MaliGAN \citep{che2017maximum} & 0.76 & 0.44 & 0.17 & 0.08 & 0.91 & 0.72 & 0.47 & \textbf{0.25} \\
LeakGAN \citep{guo2018long} & 0.84 & 0.65 & 0.44 & 0.27 & 0.94 & 0.82 & 0.67 & 0.51 \\
MLE($\alpha = 1.25^{-1}$) &  \textbf{0.93} & 0.74 & 0.51 & \textbf{0.32} & 0.93  & 0.78 & 0.59 & 0.41 \\
\hline
$\psi$-MLE($\alpha = 1.25^{-1}$) & \textbf{0.93} & \textbf{0.76} & \textbf{0.54} & \textbf{0.33} & \textbf{0.91}  & 0.75 & 0.56 & 0.38\\
\bottomrule
\end{tabular}
\end{small}
\end{center}
\label{tbl:news_bleu}
\end{table*}

\begin{table*}[!ht]
\centering
\label{tb:imagecoco}
\caption{Test BLEU and Self-BLEU on Image COCO.}
\begin{center}
\begin{small}
\begin{tabular}{l cccc | cccc }
 \toprule
 & \multicolumn{4}{c|}{BLEU($\uparrow$)} & \multicolumn{4}{c}{Self-BLEU($\downarrow$)}\\\hline 
 & 2 & 3 & 4 & 5 & 2 & 3 & 4 & 5 \\
\hline
Training Data & 0.68 & 0.47 & 0.30 & 0.19 & 0.86 & 0.62 & 0.38 & 0.42 \\
SeqGAN \citep{yu2017seqgan}  & \textbf{0.75} & 0.50 & 0.29 & 0.18 & 0.95 & 0.84 & 0.67 & 0.49 \\
MaliGAN \citep{che2017maximum} & 0.67 & 0.43 & 0.26 & 0.16 & 0.92 & 0.78 & 0.61 & 0.44 \\
LeakGAN \citep{guo2018long} & 0.74 & 0.52 & 0.33 & 0.21 & 0.93 & 0.82 & 0.66 & 0.51 \\
MLE  & 0.74 & 0.52 & 0.33 & 0.21 & \textbf{0.89} & 0.72 & 0.54 & 0.38 \\
\hline
$\psi$-MLE($\alpha = 1.0^{-1}$)  & \textbf{0.75}  &\textbf{0.53} & \textbf{0.36} &\textbf{0.23} & \textbf{0.89} & \textbf{0.70} & \textbf{0.53} &\textbf{0.36}\\\bottomrule
\end{tabular}
\end{small}
\end{center}
\label{tbl:coco_bleu}
\end{table*}

\begin{table*}[ht]
\caption{$f(t)$ and corresponding objectives($\sigma(\cdot)$ stands for sigmoid function)}
\label{tb:obj}
\centering
\begin{small}
\begin{tabular}{|c|c|}\hline 
 {$f(t)$} &\text {objectives} \\ 
\hline
{$(t-1)^{2} / 2$} & $\frac{1}{2}E_{x \sim p_{\operatorname{mix}}}(r_\psi^{2}(x)-1)-E_{x \sim p_{\text{data}}}(r_\psi(x)-1)$ \\ 
\hline {$t \log t-(1+t) \log (1+t)$} & $E_{x \sim p_{\operatorname{mix}}}(-\log(r_\psi(x)+1))-E_{x \sim p_{\text{data}}}(\log(\frac{r_\psi(x)}{r_\psi(x)+1}))$  \\ 
\hline {$\ln \left(1+e^{t}\right)$}  & $E_{x \sim p_{\operatorname{mix}}}(\sigma(r_\psi(x))r_\psi(x)-\ln \left(1+e^{r_\psi(x)}\right))-E_{x \sim p_{\text{data}}}(\sigma(r_{\psi}(x)))$ 
\\ 
\hline
\end{tabular}
\end{small}
\end{table*}
To demonstrate the effectiveness of our method, 
we conduct experiments on a synthetic setting as well as two real-world Benchmark datasets.
We compare our method with the several baseline methods, including MLE, SeqGAN \citep{yu2017seqgan}, LeakGAN \citep{guo2018long}, COT \citep{lu2018cot} and MaliGAN \citep{che2017maximum}. Note an important hyperparameter of our method is the mixture weight $m$, which is set as $\frac{1}{2}$ for default in all experiments except the ablation studies on $m$ in Sec.~\ref{abl_study}.

\subsection{Implementation Details}
\subsubsection{Bregman Divergence Minimization}
The density ratio in Sec.~\ref{DDRE} is estimated through an optimization procedure towards Bregman divergence 
A variety of functions meet the requirements of $f$, but
in all the experiments, we choose $f(t) = t \log t-(1+t) \log (1+t)$ as the default objective for its numerical stability during training. The  effect of using different objectives is also analyzed empirically in Section.~\ref{abl_study}.

\subsubsection{Variance Reduction}
The density ratio estimator, \emph{i.e.}, $r_\psi$, can be seen as the importance weight for correcting the bias in the hybrid distribution $p_{\operatorname{mix}}$. In order to increase sample quality, we apply two variance-reduction methods on importance sampling to our method \citep{mcbook,grover2019bias}:
\begin{itemize}
    \item \emph{self-normalization}: the self-normalized estimator normalizes the density ratio across a samples batch:
    \begin{equation}
    \label{eq:self-normalizaion}
    \mathrm{E}_{p_{\operatorname{mix}}(s)}[r_\psi(s)\log p_\theta(s)] \approx\sum_{i=1}^{T} \frac{r_\psi(s)}{\sum_{i=1}^{T} r_\psi(s) }\log p_\theta(s)
    \end{equation}
    where $T$ is the batch size.
    \item \emph{ratio flattening}: the density ratio can be flattened to achieve an intermediate state between the original $r_\psi(s)$ and the uniform importance weights by a parameter $\alpha \geq 0$:
    \begin{equation}
    \label{eq:ratio-flattening}
    \mathrm{E}_{p_{\operatorname{mix}}(s)}[r_\psi(s)\log p_\theta(s)] \approx\sum_{i=1}^{T} r_\psi(s)^{\alpha}\log p_\theta(s)
    \end{equation}
\end{itemize}

We find that self-normalization works best, so all experiments are implemented with self-normalization.

\subsection{Synthetic Experiments}
The synthetic experiments are conducted following the typical settings of previous works \citep{yu2017seqgan,guo2018long,lu2018cot}. 
We use a randomly initialized LSTM as the oracle model. Then we test each generation model's ability in learning from samples generated by the oracle most. 
We use a single layer LSTM with 32 hidden units. The parameters are initialized by a standard normal distribution.
With a fixed LSTM as the target, the ground-truth density is available. Hence it is possible to analyze the generation quality quantitatively by the negative log-likelihood NLL$_{oracle}$ given by the Oracle model. Besides, the log-likelihood the generative model assigns to the held-out test data, \emph{i.e.}, NLL$_{test}$ is another metric used to evaluate sample diversity. 

As is pointed out by \citep{caccia2018language}, evaluating quality alone is actually misleading for sequence generation task. 
Note the conditional probability is formalized as:
$p_\theta(s_{t}|s_{1:t-1}) = \operatorname{softmax}(o_{t}\cdot W /\alpha)$. Here $o_{t}$ is the pre-logit activation of generator, $W$ is the word embedding matrix and $\alpha$ is a Boltzmann temperature parameter. \citep{caccia2018language} introduced a temperature sweep procedure, which enumerates the possible values of $\alpha$ in a predefined range, and report the corresponding NLL$_{oracle}$ and NLL$_{test}$. In the same way, we get a curve of NLL$_{oracle}$ and NLL$_{test}$ with different temperatures (Fig.~\ref{fig:model_temp}). We find that the curve of our method is under the curves of all baseline methods, showing the superiority of our method. 

Quantitative results are reported in Table~\ref{tb:synthetic_data}, including the best NLL$_{oracle}$, NLL$_{test}$ and the comprehensive evaluation metric NLL$_{oracle}$ $+$ NLL$_{test}$. The quantitative results are obtained by tuning the temperature in the valid range defined in \citep{caccia2018language}, which indicates the quality, diversity and their trade-off of a training paradigm under the constraint that the tuned model is still a valid and information language model. Our method outperforms previous methods as it combines the strengths of MLE and GANs.
\subsection{Real Data Experiments}
We conduct  real-data experiments on two text benchmark datasets, image COCO captions and EMNLP 2017 News. 
We use BLEU score between generated samples and the whole test set to evaluate the generation quality. At the same time, we use self-BLEU \citep{zhu2018texygen} as a metric of diversity, which is the average bleu score between each generated sample and all other generated samples. Following \citep{caccia2018language}, the temperature is selected when the BLEU scores is similar to the reported numbers in \citep{guo2018long} for fair comparison.

COCO mainly contains short image captions, while EMNLP News 2017 consists of longer formal texts. The results on COCO and EMNLP News 2017 are shown in Table~\ref{tbl:coco_bleu} and Table~\ref{tbl:news_bleu} respectively. We find that our model achieves higher BLEU scores and lower self-BLEU scores, revealing better quality and higher diversity of our model.

\begin{figure}[!t]
\label{training_dy}
\centering
    \includegraphics[width=0.6 \linewidth]{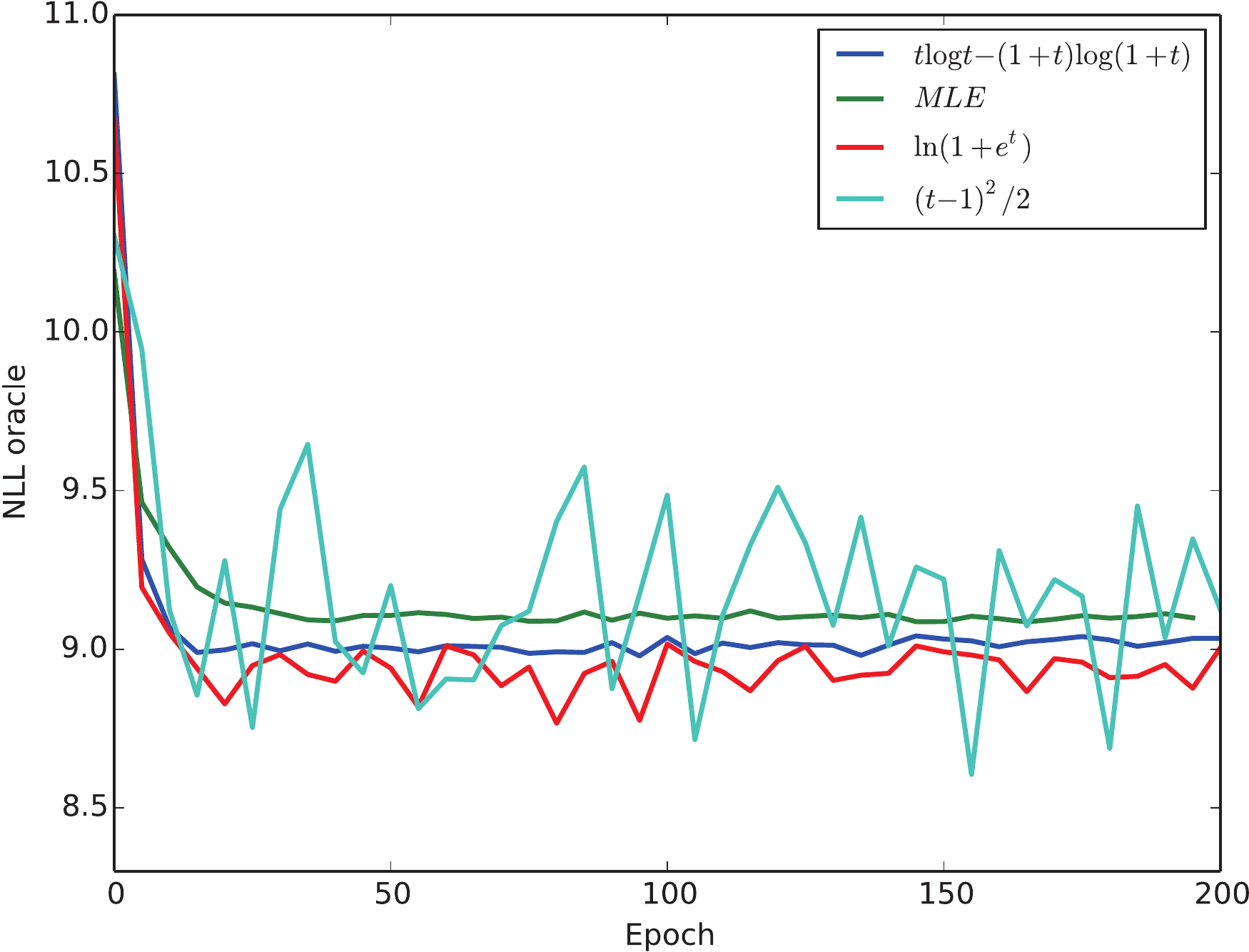}
\caption{Training dynamics of MLE and $\psi$-MLE with different $f(t)$ }
\label{fig:model_overview}
\end{figure}

\subsection{Ablation Study and Sensitive Analysis}
\label{abl_study}
\subsubsection{Mixture Weight}

One important hyperparameter in our model is the mixture weight $m$ for constructing the proposal distribution $p_{\operatorname{mix}}$.
To figure out how the method behaves with different $m$, we gradually increase $m$ from $0$ to $1$, and show the experiment results in Table~\ref{tab:an_m}. With a small $m$, it can be observed that our $\psi$-MLE has similar performance with sequential GANs. This is due to the fact that $p_{\operatorname{mix}}$ are more similar to $p_\theta$ and specifically,  our method actually degenerates to a variant of sequential GAN  when $m=0$. Correspondingly, the model is closer to MLE when $m$ approach 1. The best performance of $\psi$-MLE is achieved when $m$ is set as an intermediate value between $1$ and $0$, where both the exploration properties of GANs and the stability of MLE are incorporated. These experiments further justify the connection among $\psi$-MLE, MLE and GANs.
 
 

\begin{table}[ht]
\caption{Hyperparameter Study on Mixture Weight $m$}
\label{tab:an_m}
\begin{tabular}{|c|c|c|c|c|c|}
\hline $m$ & 0 & 1/4 & 1/2 & 3/4 & 1\\ 
\hline $NLL_\text{oracle}$ & 7.60& 6.23& \textbf{5.09}&5.63& 5.53 \\ 
\hline $NLL_\text{test}$ & 8.01& 7.91& 7.56& \textbf{7.54}& 7.60 \\ 
\hline {$NLL_\text{oracle}$} & \multirow{2}{*}{17.43}& \multirow{2}{*}{16.27} & \multirow{2}{*}{15.98}& \multirow{2}{*}{\textbf{15.94}}& \multirow{2}{*}{16.30} \\
$+NLL_\text{test}$ & & & & &\\
\hline
\end{tabular}
\end{table}

\subsubsection{Objective Density Functions}
As illustrated in Table \ref{tb:obj}, we show a family of objectives which meet the definition of Bregman Divergence and are available for direct density ratio estimation.
We conduct ablation studies within the synthetic experiment setting to find out the training dynamic of different objectives in practice. The results can be found in Fig.~\ref{fig:model_overview}. Note the training procedure with $f(t) = (t-1)^{2} / 2$  is not stable due to the numerical issues. When $f(t)$ is set as  $\ln \left(1+e^{t}\right)$ and $t \log t-(1+t) \log (1+t)$, $\psi$-MLE get remarkable improvements upon the MLE with a more stable training procedure. 

\section{Related Works}


In the context of sequence generation models, there have been fruitful lines of studies  focusing on leveraging adversarial training for sequence generation task. These works are inspired by the generative adversarial nets \citep{goodfellow2014generative}, an implicit generative model which seeks to minimize the Jensen-Shannon divergence between the generative distribution and the real data distribution through a two-play min-max game. In the sequence generation task, the gradients can not be directly back-propagated to the generative module as in continuous setting. Hence reparameterization \citep{gumbelgan} or policy gradient \citep{yu2017seqgan,guo2018long,che2017maximum} is utilized to obtain unbiased estimate of gradients. 

Our method actually can be seen as a more generalized objective family, with MLE and policy gradient based GANs as two special cases. Our method is close related to the methods leveraging tractable density distribution as noise to estimate another density, especially self-contrastive estimation~\citep{goodfellow2014distinguishability}. While self-contrastive estimation is the degenerate version of $\psi$-MLE, \emph{i.e.} directly using samples from $p_{\text{mix}}$ as ground truth to conduct MLE without the bias correction step with $r_\psi$. COT~\citep{lu2018cot} also leverages tractable density as noise. Our approach differs from COT in the calculation of the density ratio. They introduced another generative module for estimating the denominator to obtain the density ratio, while we  apply direct density ratio estimation methods which are more flexible and efficient.

Density ratio estimation has come into attention of the community of generative models. \citep{nowozin2016f} indicates a general objective family for training GANs of which the density ratio is a key element. \citep{uehara2016generative} further investigates the connection between the GANs and density ratio estimation. 
Density ratio estimator also has been utilized within the settings when the aim is to improve a learned generative model. \citep{azadi2018discriminator,turner2018metropolis} leveraged  density ratio to conduct rejection sampling over the support of generative distribution for obtaining high-quality samples. Similarly, \citep{grover2019bias} utilized an importance sampling framework to correct the biased statistics of generated distribution which result in improvements in several application scenarios of generative model.

\section{Discussion and Future Work}

We propose $\psi$-MLE, a new sequence generation training paradigm which is effective and stable when operating at large sample space encountered in sequence generation. Our method is derived based on the concept termed \emph{effect zone of supervision} which accounts for the properties of different sequence generation models. We propose a generalized family of \emph{effect zone of supervision} through self augmentation and a following density-ratio based bias correction procedure to achieve unbiased optimization during each training step. 
Experimental  results demonstrate that $\psi$-MLE is able to achieve better quality and diversity trade-off compared with previous sequence generation methods. 
An exciting avenue for future work is to extend the our training paradigm into the conditional text generation tasks,
such as machine translation, dialog system and abstractive summarization. 
Also we look forward to providing further investigation on the consistency and generalization properties of our proposed approach.

\subsubsection*{Acknowledgements}
We thank the anonymous reviewers for their insightful comments. Hao Zhou and Lei Li are the corresponding authors of this paper.

\bibliography{main}

\begin{thebibliography}{25}
\providecommand{\natexlab}[1]{#1}
\providecommand{\url}[1]{\texttt{#1}}
\expandafter\ifx\csname urlstyle\endcsname\relax
  \providecommand{\doi}[1]{doi: #1}\else
  \providecommand{\doi}{doi: \begingroup \urlstyle{rm}\Url}\fi

\bibitem[Arjovsky et~al.(2017)Arjovsky, Chintala, and
  Bottou]{arjovsky2017wasserstein}
Martin Arjovsky, Soumith Chintala, and L{\'e}on Bottou.
\newblock Wasserstein generative adversarial networks.
\newblock In \emph{International conference on machine learning}, pages
  214--223, 2017.

\bibitem[Azadi et~al.(2018)Azadi, Olsson, Darrell, Goodfellow, and
  Odena]{azadi2018discriminator}
Samaneh Azadi, Catherine Olsson, Trevor Darrell, Ian Goodfellow, and Augustus
  Odena.
\newblock Discriminator rejection sampling.
\newblock \emph{arXiv preprint arXiv:1810.06758}, 2018.

\bibitem[Caccia et~al.(2018)Caccia, Caccia, Fedus, Larochelle, Pineau, and
  Charlin]{caccia2018language}
Massimo Caccia, Lucas Caccia, William Fedus, Hugo Larochelle, Joelle Pineau,
  and Laurent Charlin.
\newblock Language gans falling short.
\newblock \emph{arXiv preprint arXiv:1811.02549}, 2018.

\bibitem[Che et~al.(2017)Che, Li, Zhang, Hjelm, Li, Song, and
  Bengio]{che2017maximum}
Tong Che, Yanran Li, Ruixiang Zhang, R~Devon Hjelm, Wenjie Li, Yangqiu Song,
  and Yoshua Bengio.
\newblock Maximum-likelihood augmented discrete generative adversarial
  networks.
\newblock \emph{arXiv preprint arXiv:1702.07983}, 2017.

\bibitem[Goodfellow et~al.(2014)Goodfellow, Pouget-Abadie, Mirza, Xu,
  Warde-Farley, Ozair, Courville, and Bengio]{goodfellow2014generative}
Ian Goodfellow, Jean Pouget-Abadie, Mehdi Mirza, Bing Xu, David Warde-Farley,
  Sherjil Ozair, Aaron Courville, and Yoshua Bengio.
\newblock Generative adversarial nets.
\newblock In \emph{Advances in neural information processing systems}, pages
  2672--2680, 2014.

\bibitem[Goodfellow(2014)]{goodfellow2014distinguishability}
Ian~J Goodfellow.
\newblock On distinguishability criteria for estimating generative models.
\newblock \emph{arXiv preprint arXiv:1412.6515}, 2014.

\bibitem[Grover et~al.(2019)Grover, Song, Agarwal, Tran, Kapoor, Horvitz, and
  Ermon]{grover2019bias}
Aditya Grover, Jiaming Song, Alekh Agarwal, Kenneth Tran, Ashish Kapoor, Eric
  Horvitz, and Stefano Ermon.
\newblock Bias correction of learned generative models using likelihood-free
  importance weighting.
\newblock \emph{arXiv preprint arXiv:1906.09531}, 2019.

\bibitem[Guo et~al.(2018)Guo, Lu, Cai, Zhang, Yu, and Wang]{guo2018long}
Jiaxian Guo, Sidi Lu, Han Cai, Weinan Zhang, Yong Yu, and Jun Wang.
\newblock Long text generation via adversarial training with leaked
  information.
\newblock In \emph{Thirty-Second AAAI Conference on Artificial Intelligence},
  2018.

\bibitem[Ho et~al.(2019)Ho, Chen, Srinivas, Duan, and Abbeel]{ho2019flow++}
Jonathan Ho, Xi~Chen, Aravind Srinivas, Yan Duan, and Pieter Abbeel.
\newblock Flow++: Improving flow-based generative models with variational
  dequantization and architecture design.
\newblock \emph{arXiv preprint arXiv:1902.00275}, 2019.

\bibitem[Kingma and Welling(2013)]{kingma2013auto}
Diederik~P Kingma and Max Welling.
\newblock Auto-encoding variational bayes.
\newblock \emph{arXiv preprint arXiv:1312.6114}, 2013.

\bibitem[Kusner and Hern{\'a}ndez-Lobato(2016)]{gumbelgan}
Matt~J Kusner and Jos{\'e}~Miguel Hern{\'a}ndez-Lobato.
\newblock Gans for sequences of discrete elements with the gumbel-softmax
  distribution.
\newblock \emph{arXiv preprint arXiv:1611.04051}, 2016.

\bibitem[Li et~al.(2016)Li, Monroe, Ritter, Galley, Gao, and
  Jurafsky]{li2016deep}
Jiwei Li, Will Monroe, Alan Ritter, Michel Galley, Jianfeng Gao, and Dan
  Jurafsky.
\newblock Deep reinforcement learning for dialogue generation.
\newblock \emph{arXiv preprint arXiv:1606.01541}, 2016.

\bibitem[Lu et~al.(2018)Lu, Yu, Zhang, and Yu]{lu2018cot}
Sidi Lu, Lantao Yu, Weinan Zhang, and Yong Yu.
\newblock Cot: Cooperative training for generative modeling of discrete data.
\newblock \emph{arXiv preprint arXiv:1804.03782}, 2018.

\bibitem[Norouzi et~al.(2016)Norouzi, Bengio, Jaitly, Schuster, Wu, Schuurmans,
  et~al.]{norouzi2016reward}
Mohammad Norouzi, Samy Bengio, Navdeep Jaitly, Mike Schuster, Yonghui Wu, Dale
  Schuurmans, et~al.
\newblock Reward augmented maximum likelihood for neural structured prediction.
\newblock In \emph{Advances In Neural Information Processing Systems}, pages
  1723--1731, 2016.

\bibitem[Nowozin et~al.(2016)Nowozin, Cseke, and Tomioka]{nowozin2016f}
Sebastian Nowozin, Botond Cseke, and Ryota Tomioka.
\newblock f-gan: Training generative neural samplers using variational
  divergence minimization.
\newblock In \emph{Advances in neural information processing systems}, pages
  271--279, 2016.

\bibitem[Owen(2013)]{mcbook}
Art~B. Owen.
\newblock \emph{Monte Carlo theory, methods and examples}.
\newblock 2013.

\bibitem[Ranzato et~al.(2015)Ranzato, Chopra, Auli, and
  Zaremba]{ranzato2015sequence}
Marc'Aurelio Ranzato, Sumit Chopra, Michael Auli, and Wojciech Zaremba.
\newblock Sequence level training with recurrent neural networks.
\newblock \emph{arXiv preprint arXiv:1511.06732}, 2015.

\bibitem[Salimans et~al.(2017)Salimans, Karpathy, Chen, and
  Kingma]{salimans2017pixelcnn++}
Tim Salimans, Andrej Karpathy, Xi~Chen, and Diederik~P Kingma.
\newblock Pixelcnn++: Improving the pixelcnn with discretized logistic mixture
  likelihood and other modifications.
\newblock \emph{arXiv preprint arXiv:1701.05517}, 2017.

\bibitem[Sugiyama et~al.(2012)Sugiyama, Suzuki, and
  Kanamori]{sugiyama2012density}
Masashi Sugiyama, Taiji Suzuki, and Takafumi Kanamori.
\newblock Density-ratio matching under the bregman divergence: a unified
  framework of density-ratio estimation.
\newblock \emph{Annals of the Institute of Statistical Mathematics},
  64\penalty0 (5):\penalty0 1009--1044, 2012.

\bibitem[Townsend et~al.(2019)Townsend, Bird, and
  Barber]{townsend2019practical}
James Townsend, Tom Bird, and David Barber.
\newblock Practical lossless compression with latent variables using bits back
  coding.
\newblock \emph{arXiv preprint arXiv:1901.04866}, 2019.

\bibitem[Turner et~al.(2018)Turner, Hung, Saatci, and
  Yosinski]{turner2018metropolis}
Ryan Turner, Jane Hung, Yunus Saatci, and Jason Yosinski.
\newblock Metropolis-hastings generative adversarial networks.
\newblock \emph{arXiv preprint arXiv:1811.11357}, 2018.

\bibitem[Uehara et~al.(2016)Uehara, Sato, Suzuki, Nakayama, and
  Matsuo]{uehara2016generative}
Masatoshi Uehara, Issei Sato, Masahiro Suzuki, Kotaro Nakayama, and Yutaka
  Matsuo.
\newblock Generative adversarial nets from a density ratio estimation
  perspective.
\newblock \emph{arXiv preprint arXiv:1610.02920}, 2016.

\bibitem[Ulyanov et~al.(2016)Ulyanov, Vedaldi, and
  Lempitsky]{ulyanov2016instance}
Dmitry Ulyanov, Andrea Vedaldi, and Victor Lempitsky.
\newblock Instance normalization: The missing ingredient for fast stylization.
\newblock \emph{arXiv preprint arXiv:1607.08022}, 2016.

\bibitem[Yu et~al.(2017)Yu, Zhang, Wang, and Yu]{yu2017seqgan}
Lantao Yu, Weinan Zhang, Jun Wang, and Yong Yu.
\newblock Seqgan: Sequence generative adversarial nets with policy gradient.
\newblock In \emph{Thirty-First AAAI Conference on Artificial Intelligence},
  2017.

\bibitem[Zhu et~al.(2018)Zhu, Lu, Zheng, Guo, Zhang, Wang, and
  Yu]{zhu2018texygen}
Yaoming Zhu, Sidi Lu, Lei Zheng, Jiaxian Guo, Weinan Zhang, Jun Wang, and Yong
  Yu.
\newblock Texygen: A benchmarking platform for text generation models.
\newblock In \emph{The 41st International ACM SIGIR Conference on Research \&
  Development in Information Retrieval}, pages 1097--1100. ACM, 2018.

\end{thebibliography}
\bibliographystyle{plainnat}

\end{document}